\documentclass[conference]{IEEEtran}
\IEEEoverridecommandlockouts

\usepackage{cite}
\usepackage{amsmath,amssymb,amsfonts}
\usepackage{algorithmic}
\usepackage{graphicx}
\usepackage{textcomp}
\usepackage{xcolor}
\def\BibTeX{{\rm B\kern-.05em{\sc i\kern-.025em b}\kern-.08em
    T\kern-.1667em\lower.7ex\hbox{E}\kern-.125emX}}

\usepackage{orcidlink}
\usepackage{cleveref}
\usepackage{siunitx}
\usepackage{booktabs}
\usepackage[inline]{enumitem}
\usepackage{balance}

\definecolor{rwthBlue}{cmyk}{1.0,0.5,0.0,0.0}
\definecolor{rwthGreen}{cmyk}{0.7,0.0,1.0,0.0}
\definecolor{rwthOrange}{cmyk}{0.0,0.4,1.0,0.0}
\definecolor{rwthRed}{cmyk}{0.15,1.0,1.0,0.0}

\Crefname{equation}{Eq.}{Eqs.}
\Crefname{figure}{Fig.}{Figs.}
\Crefname{table}{Tab.}{Tabs.}
\crefname{equation}{Eq.}{Eqs.}
\crefname{figure}{Fig.}{Figs.}
\crefname{table}{Tab.}{Tabs.}

\usepackage{pgfplots, pgfplotstable} 
\usepackage{tikz}
\pgfplotsset{compat=1.15}
\usetikzlibrary{shapes.geometric, arrows, patterns}
\usetikzlibrary{pgfplots.groupplots, matrix}
\usetikzlibrary{arrows,backgrounds}
\tikzset{every picture/.style={/utils/exec={\footnotesize}}}

\pgfplotsset{
	box plot/.style={
		/pgfplots/.cd,
		black,
		only marks,
		mark=-,
		mark size=1em,
		/pgfplots/error bars/.cd,
		y dir=plus,
		y explicit,
	},
	box plot box/.style={
		/pgfplots/error bars/draw error bar/.code 2 args={%
			\draw ##1 -- ++(1em,0pt) |- ##2 -- ++(-1em,0pt) |- ##1 -- cycle;
		},
		/pgfplots/table/.cd,
		y index=2,
		y error expr={\thisrowno{3}-\thisrowno{2}},
		/pgfplots/box plot
	},
	box plot top whisker/.style={
		/pgfplots/error bars/draw error bar/.code 2 args={%
			\pgfkeysgetvalue{/pgfplots/error bars/error mark}%
			{\pgfplotserrorbarsmark}%
			\pgfkeysgetvalue{/pgfplots/error bars/error mark options}%
			{\pgfplotserrorbarsmarkopts}%
			\path ##1 -- ##2;
		},
		/pgfplots/table/.cd,
		y index=4,
		y error expr={\thisrowno{2}-\thisrowno{4}},
		/pgfplots/box plot
	},
	box plot bottom whisker/.style={
		/pgfplots/error bars/draw error bar/.code 2 args={%
			\pgfkeysgetvalue{/pgfplots/error bars/error mark}%
			{\pgfplotserrorbarsmark}%
			\pgfkeysgetvalue{/pgfplots/error bars/error mark options}%
			{\pgfplotserrorbarsmarkopts}%
			\path ##1 -- ##2;
		},
		/pgfplots/table/.cd,
		y index=5,
		y error expr={\thisrowno{3}-\thisrowno{5}},
		/pgfplots/box plot
	},
	box plot median/.style={
		/pgfplots/box plot
	}
}

\begin{document}

\makeatletter

% \makeatletter
% \def\ps@IEEEtitlepagestyle{%
%   \def\@oddfoot{\mycopyrightnotice}%
%   \def\@evenfoot{}%
% }
% \def\mycopyrightnotice{%
%     \begin{minipage}{\textwidth}
% \centering \scriptsize
% This work has been submitted to the IEEE for possible publication. Copyright may be transferred without notice, after which this version may no longer be accessible.\hfill% <--- Change here
%     \end{minipage}
%     \gdef\mycopyrightnotice{}% just in case
% }
% \makeatother

\makeatletter
\def\ps@IEEEtitlepagestyle{%
  \def\@oddfoot{\mycopyrightnotice}%
  \def\@evenfoot{}%
}
\def\mycopyrightnotice{%
	\begin{minipage}{\textwidth}
		\centering \scriptsize
		\href{https://doi.org/10.1109/ITSC60802.2025.11423051}{DOI: 10.1109/ITSC60802.2025.11423051} \copyright 2025 IEEE. Personal use of this material is permitted. Permission from IEEE must be obtained for all other uses, in any current or future media, including reprinting/republishing this material for advertising or promotional purposes, creating new collective works, for resale or redistribution to servers or lists, or reuse of any copyrighted component of this work in other works.\hfill% <--- Change here
	\end{minipage}
	\gdef\mycopyrightnotice{}% just in case
}
\makeatother

\bstctlcite{IEEEexample:BSTcontrol}

\title{Integrated Wheel Sensor Communication using ESP32 – A Contribution towards a Digital Twin of the Road System\\
\thanks{V. Yordanov and S. Sch\"afer contributed equally to this work.}
\thanks{$^{1}$ The authors are with the Institute for Automotive Engineering, RWTH Aachen
University, {\texttt{\{prename\}.\{lastname\}@ika.rwth-aachen.de}}

$^{2}$ The authors are with the Chair of Embedded Software, RWTH Aachen
University, {\texttt{\{lastname\}@embedded.rwth-aachen.de}}

$^{3}$ B. Alrifaee is with the Department of Aerospace Engineering, University of the Bundeswehr Munich, Germany, {\texttt{bassam.alrifaee@unibw.de}}

We acknowledge the financial support for this project by the Collaborative Research Center / Transregio 339 of the German Research Foundation (DFG).
}
}

\author{
	\IEEEauthorblockN{Ventseslav Yordanov$^{1}$\,~\IEEEmembership{Student~Member,~IEEE}, Simon Schäfer$^{2}$\,\orcidlink{0000-0002-6482-2383},~\IEEEmembership{Student~Member,~IEEE}, Alexander Mann$^{2}$,\\ Stefan Kowalewski$^{2}$\,\orcidlink{0000-0002-7184-4804}, Bassam Alrifaee$^{3}$\,\orcidlink{0000-0002-5982-021X},~\IEEEmembership{Senior Member,~IEEE} and Lutz Eckstein$^{1}$\,\orcidlink{0000-0002-2586-7511},~\IEEEmembership{Senior Member,~IEEE}}
}

\maketitle

\begin{abstract}
% This paper explores a novel communication concept for the efficient transmission of integrated wheel sensor data from an ESP32 microcontroller embedded within a wheel.
% Utilizing a diverse array of sensors with varying sample frequencies, our proposed approach employs a publish-subscribe system, surpassing comparable solutions in the literature in terms of data transmission volume.
% We tested the approach, on a dynamic tire test rig, demonstrates the efficacy of the communication concept. 
% The implemented prototype sensor showcases minimal data loss, approximately 0.1\% of the sampled data, validating the reliability of the developed communication system.
% This work contributes to the advancement of real-time data acquisition, providing insights into optimizing integrated wheel sensor communication.

While current onboard state estimation methods are adequate for most driving and safety-related applications, they do not provide insights into the interaction between tires and road surfaces.
This paper explores a novel communication concept for efficiently transmitting integrated wheel sensor data from an ESP32 microcontroller. 
Our proposed approach utilizes a publish-subscribe system, surpassing comparable solutions in the literature regarding data transmission volume.
We tested this approach on a drum tire test rig with our prototype sensors system utilizing a diverse selection of sample frequencies between \SI{1}{\hertz} and \SI{32000}{\hertz} to demonstrate the efficacy of our communication concept. 
The implemented prototype sensor showcases minimal data loss, approximately \SI{0.1}{\%} of the sampled data, validating the reliability of our developed communication system.
This work contributes to advancing real-time data acquisition, providing insights into optimizing integrated wheel sensor communication.
\end{abstract}


\section{Introduction}

\subsection{Motivation}\label{sec:motivation}

Intelligent transportation systems rely on sensors to estimate their own state and that of surrounding objects. Traditional onboard methods, using inertial measurement units (IMUs), wheel encoders, and Global Navigation Satellite Systems (GNSS), operate at frequencies from 1 Hz to several hundred Hz, providing sufficient accuracy for trajectory reconstruction and stability control. However, these methods do not directly capture tire-road interactions. Even safety systems like antilock brake (ABS) and electronic stability programs (ESP) rely on conservative assumptions about forces and friction rather than real-time estimation. 

Specialized measurement vehicles used in tire testing can assess forces and moments at the tire level, allowing for the evaluation of friction-related road characteristics. However, these vehicles are costly and complex to operate.

To address this gap, we present a novel ESP32-based wireless sensor system for wheel integration. It transmits key data on tire dynamics, including inflation pressure, temperature, wheel orientation, and cavity acoustics. Due to wheel rotation, wired connections are impractical, necessitating wireless protocols. However, real-time streaming often suffers data loss during compression, potentially distorting critical acoustic signals.

Our results show that our system reliably transmits data with minimal loss, offering a cost-effective solution for real-time tire-road interaction monitoring.

\subsection{Paper Contribution}\label{sec:contribution}

This paper makes three contributions. 
\begin{enumerate*}
    \item We compare existing technologies for short-range wireless applications with our requirements of high bandwidth and low power consumption.
    \item We introduce a novel wireless communication approach for integrated wheel sensor systems.
    \item We detail experimental investigations conducted on a drum tire test rig using a prototype integrated wheel sensor, validating the efficacy of our developed communication system.
\end{enumerate*}

\subsection{Paper Structure}
\label{sec:structure}

We structure the remainder of this paper as follows: first, we present related work in \cref{sec:state_of_the_art}.
Next in \cref{sec:concept}, we describe our proposed sensor communication system while outlining the necessary software stack.
Finally in \cref{sec:results}, we discuss experimental investigations performed on a drum tire test rig under various operating conditions.

\section{State of the Art}\label{sec:state_of_the_art}

\subsection{Integrated Wheel Sensors}

Integrated wheel sensors involve attaching modules directly to the tire’s inner liner or to the rim. One of the earliest examples comes from Magori et al. \cite{MMSll}, who used a \SI{400}{\kilo \hertz} ultrasonic sensor to measure the distance between the liner and the rim, relying on a wired A/D–D/A interface. Matsuzaki et al. \cite{MT05} introduced a passive wireless approach based on changes in an LC circuit’s resonant frequency to capture carcass deformation. Later, they reviewed a variety of “intelligent tire” sensors but gave only brief attention to communication methods \cite{MT08}. Hasan et al. \cite{HAH+11} developed a tire pressure monitoring system that transmits at a rate of \SI{0.125}{\hertz} using a custom \SI{10}{\kilo {bit} \per \second} radio. Masino et al. \cite{MAS16,MAS17a} created a system to capture in-tire acoustic data at \SI{5}{\kilo \hertz}, moving that data first via a wired link from the acoustic sensor to an external circuit board, then sending it wirelessly over Bluetooth to an in-cabin recorder. Pozuelo et al. \cite{GPOY+17} deployed wired strain sensors (sampled at \SI{5}{\kilo \hertz}) to measure tire deflections, routing cables through additional rim valves.

A significant body of work applies accelerometers to the tire’s inner liner. Savaresi et al. \cite{STLD08} estimated road–tire contact forces by wirelessly transmitting accelerometer data with a Datatel telemetry system, sampling auxiliary Hall sensors at \SI{10}{\kilo \hertz}. Ergen et al. \cite{ESVS+09} leveraged a custom ultra-wideband (UWB) radio (\SI{>2}{\mega {bit} \per \second}) for computing tire kinematics, organizing multiple wheel sensors into a cluster-tree network with a time-slotted MAC protocol and a central system control host. Hong et al. \cite{HEHB13} measured tire lateral deflection, synchronizing each wheel’s accelerations with vehicle CAN-based data, including GPS and IMU signals. Finally, Lee et al. \cite{LKKL21} sampled wheel accelerations at \SI{1}{\kilo \hertz} and transferred them via Bluetooth to a telemetry device on the outer wheel, primarily to gauge road surface conditions. 

% Collectively, these approaches balance factors like sampling rate, bandwidth availability, communication architecture, and power usage, reflecting diverse strategies for meeting real-time constraints with integrated wheel sensing systems.

\subsection{Sensor Network Communications}

Connecting multiple sensors in a network and making their data accessible to various data storage systems has been widely discussed in numerous publications. This includes research on Internet of Things (IoT) networks, which are defined by numerous low-powered devices and sensors designed to operate in an energy-efficient manner.

Much work has compared wireless technologies for short-range communication. For instance, Cho et al. \cite{CPH+14} analyzed BLE’s latency, showing its potential for real-time, low-power connections. Danbatta et al. \cite{DV19} contrasted ZigBee, Z-Wave, Wi-Fi, and Bluetooth in smart home and IoT scenarios, noting that each excels in different factors such as cost, ease of use, range, and scalability. Dizdarevic et al. \cite{DCJMB19} shifted focus to application-layer protocols, highlighting MQTT’s stability and CoAP’s emerging capabilities, yet concluding that no single protocol meets all needs.

For extended-range applications, Mekki et al. \cite{MBCM18} evaluated low-power WANs like LoRaWAN, Sigfox, and NB-IoT, finding them well-suited for infrequent data transmission but typically limited in throughput. Moving to real-time performance, Cosar et al. \cite{CMB11} introduced A-Stack, a real-time protocol for 802.15.4 multi-hop networks, while Kim et al. \cite{KPK+17} surveyed broader approaches for handling soft, hard, and firm real-time constraints, addressing routing algorithms and MAC-layer strategies. In Wi-Fi-based solutions, Wei et al. \cite{WLH+13, WLC+18} introduced RT-WiFi, which uses TDMA for sampling frequencies up to \SI{6}{\kilo \hertz}, though power management was initially underemphasized.

Meanwhile, DDS emerged for real-time, publish-subscribe systems. Although commonly thought unsuitable for low-power micro-controllers, several adaptations exist. González et al. \cite{GMV+11} proposed $\mu$DDS but evaluated it mainly on mid-range processors, leaving open questions for ultra-low-power platforms. Two open-source implementations address the gap further: XRCE-DDS \cite{DGK+21} employs a client-agent pattern to reduce overhead, and EmbeddedRTPS \cite{KWAK19} uses a fully distributed approach, enabling micro-controllers to directly join DDS networks without intermediaries.

\section{Development of a Real-Time Communication for Integrated Wheel Sensor}
\label{sec:concept}
Designing a communication concept requires balancing multiple factors, with no single best approach. This work focuses on three key aspects: selecting the appropriate communication technology and protocol stack, designing an effective network architecture, and creating an real-time capable firmware for the sensor system.

Network architecture depends on the chosen communication technology, as its characteristics directly impact system design. For example, using short-range technologies like Bluetooth for wheel sensor data may require nearby intermediaries to ensure reliable transmission, potentially limiting scalability.

Selecting an appropriate communication technology and protocol stack requires a structured approach. The process begins with defining the system’s communication requirements, followed by evaluating against available technologies, primarily focusing on the physical layer standards \cite{KDD14}. After narrowing down the options, compatible application-layer protocols suitable for micro-controllers are assessed. The optimal combination of physical and application layers is then selected, ensuring alignment with system constraints and performance needs.

Intermediate layers are generally not considered separately, as they are determined by the selected stack. For example, DDS over Wi-Fi typically relies on TCP or UDP with IP, while Bluetooth configurations depend on data type and timing constraints. These dependencies highlight the importance of an integrated approach when designing a communication system.

\subsection{Requirements}\label{sec:requirements}

For the reference and testing implementation of this work, a prototype integrated wheel sensor, shown in \cref{fig:estp}, developed at the Institute for Automotive Engineering (ika) RWTH Aachen University is used. The wheel sensor consist of an acoustic module (AM), an inertial measurement unit (IMU) with six degrees of freedom (DoF), and a combined temperature and pressure module (TP). All three sensor modules are based on Micro-Electromechanical Systems (MEMS) technology. The data acquisition and communication tasks are performed on a dual-core ESP32 micro-controller. The wheel sensor is powered by an integrated lithium battery. A voltage divider is used to monitor the battery state of charge (BSoC). Details on the precision, sample rate, and resulting data rate of the individual sensor components are provided in \cref{tab:sdchar}.

% The requirements of the communication concept are derived from the use-case of the integrated wheel sensor as described in previous work \cite{YOR23,YOR25,GBO25} and are given in \cref{tab:req}. 

The requirements of the communication concept are derived from the use-case of the integrated wheel sensor as described in \cite{YOR23,YOR25,GBO25}. Since the primary function of the integrated wheel sensor is to provide input data for a real-time tire simulation model, meeting real-time and reliable communication requirements are key. The sensor data are used to determine the road surface characteristics affecting the vehicle and the road. This data needs to be received in a consistent and predictable manner.

The fourth requirement is a data rate of at least 1078 Kbit/s, based on the data rates provided in \cref{tab:sdchar}.

\begin{table}[t!]
    \caption{Sensor data characteristics}
    \label{tab:sdchar}
    \centering
    \begin{tabular}{|l|l|l|l|}
        \toprule
        Sensor & Precision & Sample rate & Data rate\\
        \midrule
        AM (VM2020) & \SI{32}{{bit}} & \SI{32000}{\hertz} & \SI{1024}{\kilo {bit}\per\second}\\
        IMU (ICM-20649) & \SI{16}{{bit}} & \SI{562.5}{\hertz} & \SI{54}{\kilo {bit}\per\second}\\
        TP (MS5803-14BA) & \SI{32}{{bit}} & \SI{5}{\hertz} & \SI{0.32}{\kilo {bit}\per\second}\\
        BSoC & \SI{12}{{bit}} & \SI{1}{\hertz} & \SI{0.012}{\kilo {bit}\per\second}\\
        \bottomrule
    \end{tabular}
\end{table}

Additional requirements for the communication technology pertain to the range and number of participants (nodes). The minimum range required to transmit data from a wheel sensor to the receiving device, located at least on the fender liner, should be more than one meter. The integrated wheel sensor concept is designed to be used on each wheel of a vehicle. For a typical passenger car, this results in at least four sensors plus the computer to process the data. For commercial vehicles, this system should scale to accommodate eight or more wheels.

\subsection{Selection of Communication Technology}\label{sec:selection_technology}

The following overview covers wireless technologies used for communication between multiple devices. This list includes publicly available technologies with established standards. To select the most suitable communication technology based on the requirements, an overview is provided in \cref{tab:ctover}.

\begin{table}[t!]
    \caption{Communication technologies overview}
    \label{tab:ctover}
    \centering
    \begin{tabular}{|l|l|l|l|l|}
    \toprule
        Name & Latency & Data rate & Range & Nodes\\ 
        \midrule
        \textbf{Wi-Fi} \cite{TVLZ15} & \SI{0.6}{\milli \second}    & \SI{600}{\mega {bit}\per\second} & \SI{100}{\meter} & 32\\
        BL \cite{CPTT18} & \SI{3}{\milli \second}      & \SI{2}{\mega {bit}\per\second} & \SI{40}{\meter} & 7\\
        UWB \cite{MO18} & \SI{>1}{\milli \second}     & \SI{600}{\mega {bit}\per\second} & \SI{>10}{\meter} & High\\
        BLE \cite{RGL17} & \SI{3}{\milli \second}      & \SI{1}{\mega {bit}\per\second} & \SI{10}{\meter} & 7\\
        LR-WPAN \cite{CPTT18} & \SI{4}{\milli \second}      & \SI{0.25}{\mega {bit}\per\second} & \SI{150}{\meter} & High\\
        Z-Wave \cite{DV19, YMK16} & \SI{20}{\milli \second}     & \SI{0.1}{\mega {bit}\per\second} & \SI{40}{\meter} & 200\\
        LoRa \cite{SWH17} & \SI{>100}{\milli \second}   & \SI{0.05}{\mega {bit}\per\second} & \SI{>1}{\kilo\meter} & 5\\
        LTE-M \cite{DV20} & \SI{6}{\second}   & \SI{4}{\mega {bit}\per\second} & \SI{>1}{\kilo\meter} & 60000\\
        NB-IoT \cite{SWH17} & \SI{10}{\milli \second}     & \SI{0.25}{\mega {bit}\per\second} & \SI{>1}{\kilo\meter} & High\\
        \bottomrule
    \end{tabular}
\end{table}

\textbf{Wi-Fi (IEEE 802.11x)} covers various 802.11 revisions (e.g., n, ac, ax) that operate in the \SI{2.4}{\giga \hertz} and \SI{5.0}{\giga \hertz} bands. Each newer standard integrates improvements in throughput, latency, and coverage. For instance, 802.11n (Wi-Fi 4) reaches data rates of up to \SI{600}{\mega {bit} \per \second}, suitable for real-time tasks if interference is minimal. Indoors, Wi-Fi typically offers robust connections within \SI{100}{\meter}, but outdoor ranges can extend much further in open space. Latency can remain below \SI{1}{\milli \second} in controlled environments, though interference may increase it slightly. Up to 32 devices can be supported per router, though this depends on hardware.

\textbf{Bluetooth (BL)} likewise operates at \SI{2.4}{\giga \hertz}, with Bluetooth 5 improving on older versions in terms of coverage (up to \SI{\sim40}{\meter} indoors and \SI{200}{\meter} outdoors), data rates of up to \SI{2}{\mega {bit} \per \second}, and latencies under \SI{3}{\milli \second}. Battery consumption also decreased in recent revisions. While there is no strict limitation on connections, typical devices often allow up to seven simultaneous links. \textbf{Bluetooth Low Energy (BLE)}, introduced with Bluetooth 4.x, further cuts power consumption, at the cost of halving data rates to about \SI{1}{\mega {bit} \per \second}, making it fit for scenarios where minimal energy use and moderate data throughput are required. BLE achieves latency around \SI{3}{\milli \second}, and has an approximate range of \SI{10}{\meter} which can be sufficient for inter-vehicle sensor applications given an appropriate network structure.

\textbf{Ultra-Wideband (UWB)} transmits data using very large bandwidth channels (\SI{\geq500}{\mega \hertz}), often at short range (up to \SI{\sim10}{\meter}) with rates above \SI{600}{\mega {bit} \per \second}. The UWB physical layer can also be configured for long-range communication up to \SI{100}{\meter}, though this reduces the data rate. Its physical layer enable precise localization, making UWB a popular choice for localization applications. Since UWB uses radio frequencies similar to Wi-Fi and Bluetooth and operates over short distances, it can also provide low-latency data transfer. However, mature real-time capable implementations and standardized hardware solutions are still developing, limiting its widespread adoption beyond localization use cases.

Two other standards, \textbf{IEEE 802.15.4 (LR-WPAN)} and \textbf{Z-Wave}, target low-power, low-data-rate applications. \textbf{LR-WPAN} supports data rates up to \SI{250}{\kilo {bit} \per \second} with ranges up to \SI{\sim150}{\meter}. Mesh networking and multi-hop features allow large-scale sensor networks with minimal power usage. Latency is approximately \SI{4}{\milli \second}, and devices can use underutilized frequency bands below \SI{1.0}{\giga \hertz} to reduce interference. \textbf{Z-Wave}, which operates from \SI{\sim850}{\mega \hertz} to \SI{950}{\mega \hertz}, achieves data rates near \SI{100}{\kilo {bit} \per \second} with typical indoor ranges of \SI{\sim40}{\meter}. Networks can include hundreds of nodes, making it popular in home automation contexts.

For long-range but low-throughput needs, \textbf{LoRa} can cover multiple kilometers. However, its data rate of around \SI{50}{\kilo {bit} \per \second} and relatively high latency, due to long sleep phases for power savings, render it impractical for time-sensitive tasks. Meanwhile, \textbf{LTE-M} and \textbf{NB-IoT} extend cellular standards (3GPP releases) to handle IoT-specific demands. \textbf{LTE-M} can deliver data rates up to \SI{4}{\mega {bit} \per \second}, with latency dropping to milliseconds under ideal conditions but reaching seconds in poor signal situations. \textbf{NB-IoT} focuses on power saving and scalability at the cost of lower throughput (around \SI{250}{\kilo {bit} \per \second}).

Overall, three technologies exhibit promising characteristics for consideration in this work: Bluetooth, UWB and Wi-Fi. However, due to UWB's lack of standardized components and limited research on real-time applicability, it has been excluded from further consideration. Choosing between Bluetooth and Wi-Fi primarily hinges on their respective promises. Both technologies meet the requirements, but Wi-Fi stands out for its higher data rate. Moreover, Wi-Fi offers a simpler architecture that scales effectively with varying numbers of wheel sensors. In contrast, Bluetooth would likely necessitate intermediaries between wheel sensors and the computer, adding complexity and cost to the setup. Given these factors, we selected Wi-Fi as the preferred technology for our application.

\subsection{Selection of Application Protocol}\label{sec:selection_protocol}

This section compares various application protocols for real-time communication. \Cref{tab:protocols} offers an overview of their typical latency, communication paradigm, and whether they require a centralized coordinator.

\begin{table}[t!]
    \caption{Application protocols overview}
    \label{tab:protocols}
    \centering
    \begin{tabular}{|l|l|l|l|}
        \toprule
        Name & Latency & Paradigm  & Coordinator\\ 
        \midrule
        AMQP \cite{LIN17} & \SI{360}{\milli \second} & Pub/Sub & Yes\\
        CoAP \cite{LMC13} & \SI{400}{\milli \second} & Req/Resp & No\\
        \textbf{DDS} \cite{DGK+21, KWAK19, GMV+11}& \SI{2}{\milli \second} & Pub/Sub & No\\
        MQTT \cite{LIN17} & \SI{130}{\milli \second} & Pub/Sub & Yes\\
        XMPP \cite{PKBT18} & \SI{600}{\milli \second} & Pub/Sub & Yes\\
        \bottomrule
    \end{tabular}
\end{table}

% \textbf{Advanced Message Queuing Protocol} (AMQP) is an open standard designed for application layer network protocols, utilizing message-oriented middleware. It operates over a reliable transport layer protocol such as TCP and supports both client/server and publish/subscribe communication models. AMQP includes features such as flow control, various levels of Quality of Service (QoS), and user authentication and encryption. In latency tests conducted by Linden et al., AMQP demonstrated an average latency of 111 ms for small payloads of a few bytes. However, as the payload size increases, the average latency rises to approximately 360 ms at 250 KBytes .

\textbf{Advanced Message Queuing Protocol (AMQP)} employs message-oriented middleware and supports both client/server and publish/subscribe models. Tests show it can process small payloads in about \SI{111}{\milli \second} but rises to \SI{\sim360}{\milli \second} for \SI{250}{\kilo {byte}} payloads \cite{LIN17}. It includes flow control, multiple Quality of Service (QoS) levels, encryption, and authentication.

% \textbf{Constrained Application Protocol} (CoAP) is an application layer protocol designed for resource-constrained devices to connect to the Internet. It offers Representational State Transfer (RESTful) web services as an alternative to Hypertext Transfer Protocol (HTTP) and uses UDP as its transport protocol, while still enabling reliability over UDP if needed. CoAP provides various levels of security. Testing an implementation of CoAP on resource-constrained devices demonstrated latency of 400 ms or less, depending on the payload size \cite{LMC13}.

\textbf{Constrained Application Protocol (CoAP)} is a RESTful protocol running mainly over UDP with low resource demands. Though it can achieve latency of \SI{\sim400}{\milli \second} under specific test conditions \cite{LMC13}, this is comparatively high for real-time needs. CoAP can provide reliable communication atop UDP if necessary, with security features as well.

\textbf{Data Distribution Service (DDS)} is a real-time capable publish-subscribe middleware standard from the Object Management Group (OMG). It handles data-centric communication and supports self-discovery of network nodes, removing the need for a central broker. DDS also defines multiple QoS settings, from best-effort to fully reliable modes, and uses the Real-Time Publish-Subscribe (RTPS) protocol for its transport layer. Open-source implementations like EmbeddedRTPS allow microcontrollers to operate as first-class DDS participants with latencies around \SI{2}{\milli \second} for \SI{1024}{{byte}} payloads \cite{KWAK19}.

% \textbf{Message Queuing Telemetry Transport} (MQTT) is a lightweight publish-subscribe network protocol specifically designed for resource-constrained devices. It operates over an underlying transport protocol that is ordered, lossless, and bi-directional, such as TCP. MQTT facilitates publish/subscribe communication through two main entities: message brokers and clients. Clients are the communicating devices connected to a broker, while the message broker is a server responsible for coordinating the network. The broker receives messages from clients and routes them to their destinations. MQTT supports various QoS levels, including the retention of messages when there are no subscribers for a specific topic. Messages are organized into topics, allowing devices to publish messages to a topic and subscribe to a topic to receive information. The broker maintains a list of subscribers for each topic and ensures the delivery of published messages to them. Latency tests by \cite{LIN17} show that MQTT achieves reliable communication with latency's of less than 130 ms for payload sizes up to 250 KBytes.

\textbf{Message Queuing Telemetry Transport (MQTT)} is a lightweight publish-subscribe protocol that relies on a central broker for message routing. Designed for constrained devices, MQTT can achieve latencies of \SI{\sim130}{\milli \second} for \SI{250}{\kilo {byte}} messages, offering different QoS levels, topic-based messaging, and message retention if no subscriber is currently online \cite{LIN17}.

% \textbf{Extensible Messaging and Presence Protocol} (XMPP) is an open standard developed by the Internet Engineering Task Force (IETF) that enables near real-time data exchange between multiple network entities. It supports various applications, including Voice over Internet Protocol (VoIP), instant messaging, and video transfer. XMPP utilizes a decentralized client-server model, meaning there is no central server responsible for coordination, and clients are identified by unique XMPP addresses. While XMPP relies on TCP for reliable communication, it does not offer additional QoS levels. The protocol includes various security options, such as point-to-point security. Tests by Pohl et al. reveal that XMPP exhibits high latency, starting at approximately 600 ms and increasing to nearly 10 sec as payload size grows \cite{PKBT18}.

\textbf{Extensible Messaging and Presence Protocol (XMPP)} uses a decentralized client-server model for near real-time data exchange. While suitable for voice and video, tests indicate it has a minimum latency of \SI{\sim600}{\milli \second} that grows with payload size, which disqualifies it for strict real-time scenarios \cite{PKBT18}.

% An overview of the discussed application protocols with respect to the communication requirements is provided in \cref{tab:protocols}. XMPP has been found to have high latency in published studies, which is why it is not being considered further.
% Additionally, a protocol that supports publish-subscribe operations is preferred due to its ability to scale efficiently to a large number of sensors and its superior support for many-to-one and one-to-many communication. Moreover, CoAP is offers the second highest latency and thus has been excluded from the final selection.
% Among the remaining protocols (AMQP, DDS, and MQTT), DDS stands out as the only protocol with self-discovery capabilities, eliminating the need for additional server coordination. Reliable communication between the micro-controller and the connected computer is crucial, and introducing an intermediary could become a potential source of errors without significantly reducing the micro-controller's burden of re-sending erroneous packets. Therefore, DDS is selected as the preferred application layer protocol. Specifically the EmbeddedRTPS implementation, is utilized because it is open-source with outstanding latency.

As \cref{tab:protocols} indicates, XMPP has the highest minimum latency and is thus excluded. CoAP also has relatively high latency (\SI{\sim400}{\milli \second}). Among the remaining protocols (AMQP, DDS, MQTT), DDS stands out for its self-discovery and direct peer-to-peer capability, reducing reliance on a central coordinator. Its QoS features can ensure reliable transmission without an intermediary. Therefore, DDS is selected as the preferred application protocol, specifically using the open-source EmbeddedRTPS implementation, which offers excellent real-time performance.

\subsection{Implementation}

Due to the real-time requirements for sensor data communication, the chosen operating system (OS) must be real-time capable and support ESP32. FreeRTOS, a widely adopted OS for micro-controllers, is well-maintained for ESP32. Espressif offers an adapted version of FreeRTOS that supports various ESP32 hardware features, including dual-core functionality. Time in FreeRTOS is measured in ticks, configurable between \SI{100}{\hertz} and \SI{1000}{\hertz}, with a higher tick rate providing better time resolution at the cost of increased number of time handling interrupts. For this work, we set it to \SI{1000}{\hertz} for \SI{1}{\milli \second} resolution and quick context switching for critical tasks.

Tasks in FreeRTOS can be in Running, Ready, Blocked, or Suspended states, with state transitions. The OS uses a fixed priority preemptive scheduler with round-robin time-slicing for tasks of equal priority, where each time-slice corresponds to a single tick interrupt. The ESP32 version includes core affinity for assigning tasks to specific cores. FreeRTOS's IDLE task, the lowest priority task, triggers the scheduler when no other tasks are runnable. For inter-task and inter-core communication, FreeRTOS provides event bits and queues. Event bits signal events and can be grouped into event groups to avoid race conditions, while queues, which are thread-safe FIFO structures, facilitate communication without data overwriting issues. Additionally, FreeRTOS includes mutexes to prevent deadlocks.

In this work we combine FreeRTOS with the Robot Operating System (ROS) for access to additional tooling and functionality. We implement Micro-ROS as a tailored version of ROS designed for low-end devices, providing real-time operability and interoperability with ROS. It supports most ROS2 features but has restrictions like fixed-size messages to avoid dynamic memory allocation. Micro-ROS runs on FreeRTOS  and allows developers to control the order of ROS entity operations. It also supports EmbeddedRTPS as middleware. 

EmbeddedRTPS allows to configure some QoS settings and constrain many objects
like e.g. endpoints running on the device to limit the amount of memory that is used.
Because reliable communicators are required, each of the RTPSWriters here will hold
proxies of connected RTPSReaders, which further increases memory requirements
for each endpoint. Furthermore, due to the high sample rate of especially the AM
and the IMU and the limited memory of the ESP32, data is required to be
published at a relatively high frequency. As multiple reliable endpoints with high
frequency rate would need to respond to ACKNACKS from connected readers and
hold a history of their messages, one message is implemented to hold all the sensor
data.  For network connectivity, the Espressif version of FreeRTOS relies on lightweight IP (LwIP).

To optimally utilize the dual-core feature of the micro-controller, we split the program tasks between the two cores. The first core is performing the Wi-Fi and Command tasks, while the second core is performing the Micro-ROS and data acquisition tasks. Due to the high difference in the sensor sample rates (see \cref{tab:sdchar}), we implement separate data acquisition tasks for each sensor module. The tasks with the corresponding priority and task periodicity are provided in \cref{tab:tsche}. To monitor the performance of the wireless communication, the message count and the time stamp for each sensor task are transmitted.

\begin{table}[t!]
    \caption{Task schedule}
    \label{tab:tsche}
    \centering
    \begin{tabular}{|l|l|l|l|}
    \toprule
        Core 1 & Core 2 & Priority & Periodicity\\
        \midrule
        Wi-Fi Task & Micro-ROS Task & 1 & \SI{4}{\milli \second}\\
        Command Task & AM Task & 2 & \SI{7}{\milli \second}\\
         & IMU Task & 3 & \SI{10}{\milli \second}\\
         & TP Task & 4 & \SI{200}{\milli \second}\\
         & BSoC Task & 5 & \SI{1}{ \second}\\
         \bottomrule
    \end{tabular}
\end{table}

\section{Experimental Investigations}\label{sec:results}

\begin{figure}[t!]
    \centering
    \includegraphics[width=1.0\linewidth]{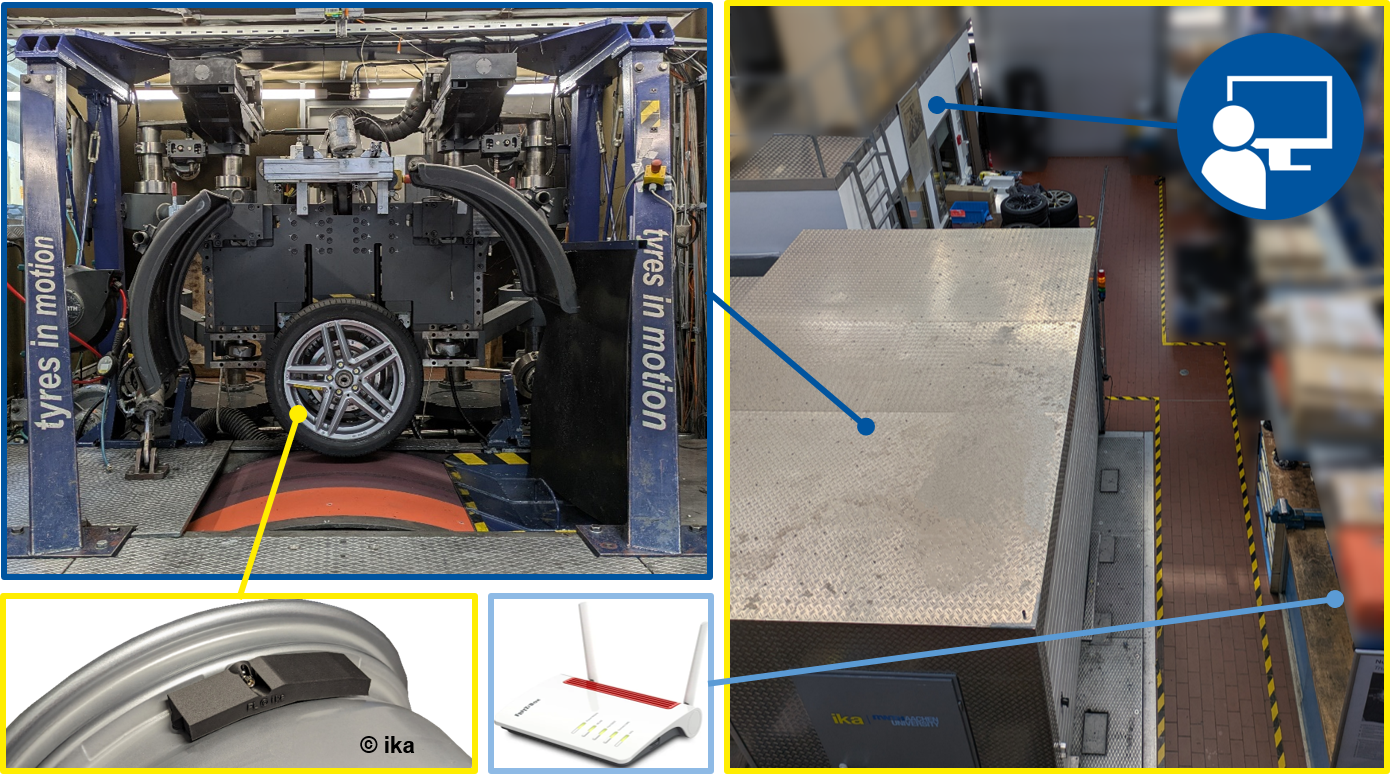}
    \caption{The experimental setup demonstrates the tire test rig (top left) communicating with the router outside its metal enclosure (right). The enclosure simulates the vehicle’s metal body.}
    \label{fig:estp}
\end{figure}

To investigate the performance of the developed sensor communication, tire measurements were conducted on a drum tire test rig. This allows simulation of realistic vehicle operating conditions in the laboratory environment. The precise control of velocity and wheel load offers the possibility of assessing the robustness of the wireless communication in various scenarios. The experimental setup is shown in \cref{fig:estp}. A prototype of the integrated wheel sensor was attached to a 17-inch wheel through the pressure valve within the air cavity of the tire. The tire used in this investigation is Pirelli Powergy with dimensions 225/45R17, load index 94 and speed index Y. To simulate the indirect propagation of Wi-Fi waves in a vehicle, the network router was placed outside the closed test rig. The receiver computer was located in the test rig control room. The receiver software is implemented in a MATLAB application using the ROS Toolbox. During recording, the received sensor messages are cached in RAM of the computer and stored on the hard drive as rosbag files.

\begin{figure}[t!]
    \centering
    \includegraphics[width=1.0\linewidth]{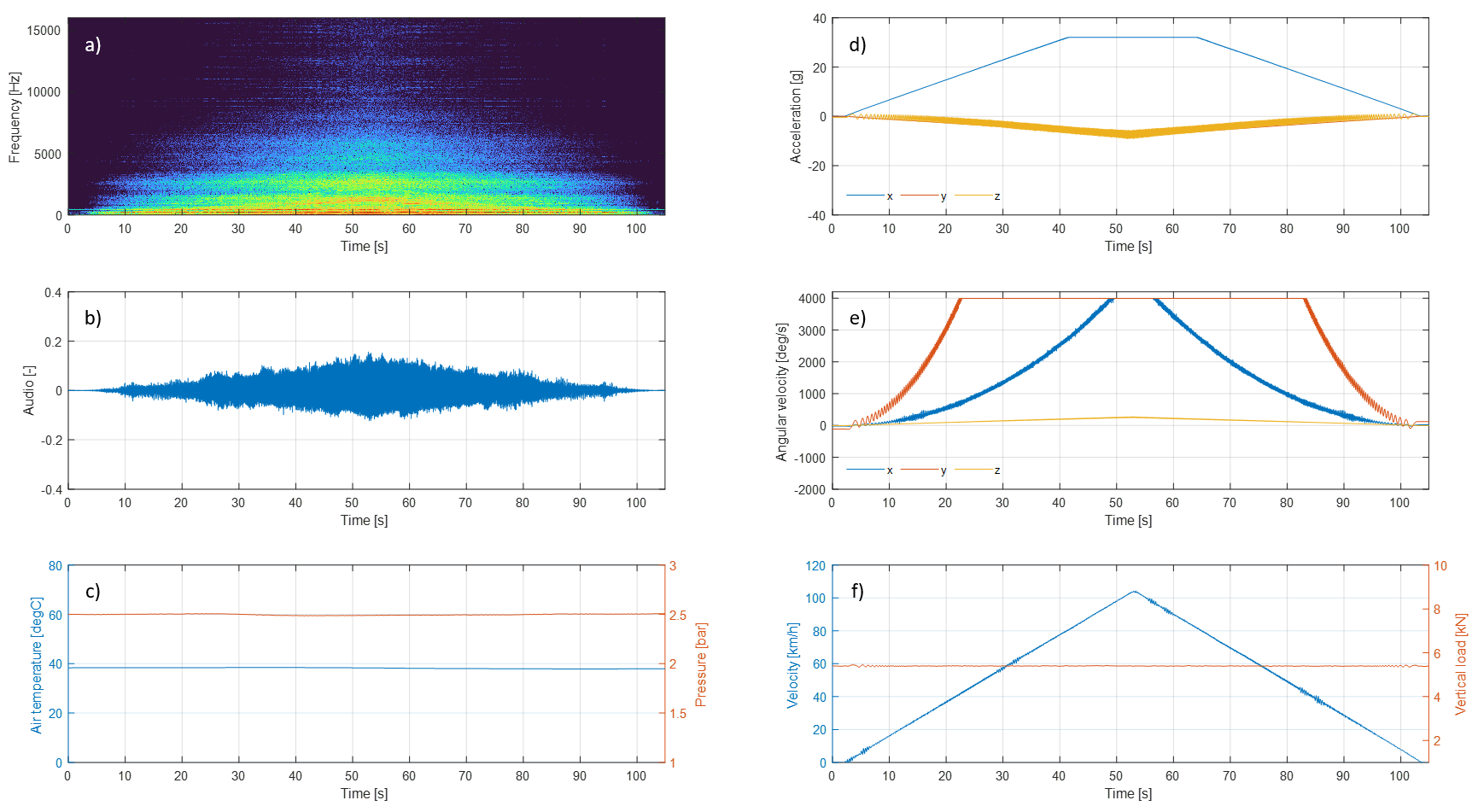}
    \caption{Experiment data - Vertical load: 5256 [N], Inflation pressure: 2.5 [bar] a) Spectrogram of audio data, b) Sensor audio data, c) Sensor temperature and pressure data, d) Sensor acceleration data, e) Sensor gyroscope data, f) Test rig velocity and vertical load data.}
    \label{fig:sde}
\end{figure}

With this setup, four measurements were made, in which the tire was inflated with 2.5 bar and accelerated from stand still to a rolling velocity of 100 km/h. For each of the four measurements, the vertical load applied to the wheel was held constant at 2628 N, 5256 N, 7884 N and 9198 N. The selected values are based as a percentage proportion of the tire load index and represent vertical loads typical for dynamic vehicle maneuvers. An example of the experiment data for the second vertical load is provided in \cref{fig:sde}.

\begin{table}[t!]
    \centering
    \caption{Communication performance}
    \label{tab:tcmp}
    \begin{tabular}{|l|l|l|l|l|l|}
    \toprule
        Sensor & Received & Expected & Mean & Minimum & Maximum\\ 
         & messages & gap [ms] & gap [ms] & gap [ms] & gap [ms]\\
         \midrule
        AM & 59318 & 7 & 6.999 & 4.997 & 8.998\\
        IMU & 41521 & 10 & 10.001 & 7.406 & 12.649\\
        TP & 2077 & 200 & 199.999 & 197.438 & 202.623\\
        BSoC & 415 & 1000 & 1000.001 & 998.646 & 1001.469\\
        \bottomrule
    \end{tabular}
\end{table}

Using the message counts and time stamps recorded during the measurements, the communication performance in terms of the periodicity of the messages was assessed. The results for each of the tasks are presented in \cref{tab:tcmp}. The calculated mean gap between sensor tasks ideally matches the corresponding expected gap, highlighting the fulfillment of the real-time and latency requirements. During the measurements no message losses occurred, also fulfilling the reliability requirement. It is worth noting that during investigations where received data was stored directly on the hard drive, message losses of up to \SI{0.1}{\%} were observed. This may be related to hard drive speed; however, the authors cannot confirm this at this time.

\section{Conclusion}\label{sec:conclusion}

In this paper, we compared existing short-range wireless technologies against the requirements of an integrated wheel sensor system. Based on this analysis, we developed a novel wireless communication approach and implemented it on an ESP32 micro-controller. We evaluated its performance using a prototype wheel sensor on a drum tire test rig, simulating dynamic vehicle conditions. The results confirmed that our approach meets the requirements, demonstrating high reliability and low latency.

Possible future research should evaluate the security and scalability of the presented approach. While the publish-subscribe approach employed by ROS allow for easy extension to multiple sensors, a detailed analysis of the resulting traffic and potential bandwidth limitations is necessary. Additionally, the security of this approach should be assessed. The current implementation relies on the Wi-Fi network to handle security, which could open a possible attack corridor. However, additional validation by the network router and possible encryption could add to the processing pipeline and potentially violate the real-time requirement.

\bibliographystyle{IEEEtran}
\bibliography{library.bib}
% \balance

\end{document}